\documentclass[11pt,a4paper]{article}
\usepackage[hyperref]{naaclhlt2019}
\usepackage{times}
\usepackage{latexsym}

\usepackage{url}
\usepackage{linguex}
\usepackage{booktabs}
\usepackage{cancel}
\usepackage{amsmath}
\usepackage{amssymb}
\usepackage{mathtools}
\usepackage{bm}
\usepackage{centernot}
\usepackage{multirow}
\usepackage[normalem]{ulem}

\usepackage{tikz}
\usepackage{tikz-dependency}

\aclfinalcopy 
\def\aclpaperid{***} 

\title{The lexical and grammatical sources of neg-raising inferences}

\author{Hannah Youngeun An \\
        Department of Computer Science\\
        University of Rochester\\\And
        Aaron Steven White \\
        Department of Linguistics\\
        University of Rochester \\}

\date{}

\begin{document}
\maketitle
\begin{abstract}
We investigate \textit{neg(ation)-raising} inferences, wherein negation on a predicate can be interpreted as though in that predicate's subordinate clause. To do this, we collect a large-scale dataset of neg-raising judgments for effectively all English clause-embedding verbs and develop a model to jointly induce the semantic types of verbs and their subordinate clauses and the relationship of these types to neg-raising inferences. We find that some neg-raising inferences are attributable to properties of particular predicates, while others are attributable to subordinate clause structure.
\end{abstract}

\setlength{\Exlabelsep}{0em}%
\setlength{\SubExleftmargin}{1.2em}%
\setlength{\Extopsep}{.3\baselineskip}%

\renewcommand{\firstrefdash}{}

\section{Introduction}
\label{sec:introduction}

Inferences that are triggered (at least in part) by particular lexical items provide a rich test bed for distinguishing the relative semantic contribution of lexical items and functional structure. One class of such inferences that has garnered extended attention is \textit{neg(ation)-raising}, wherein negation on a predicate can be interpreted as though in that predicate's subordinate clause \citep{fillmore-1963, bartsch-1973, horn-1978, gajewski-2007}. For example, a neg-raising inference is triggered by \ref{ex:neg-raising} while one is not triggered by \ref{ex:non-neg-raising}.

\ex. \label{ex:neg-raising} Jo doesn't think that Bo left.\\$\rightsquigarrow$Jo thinks that Bo didn't leave.

\ex. \label{ex:non-neg-raising} Jo doesn't know that Bo left.\\$\not\rightsquigarrow$Jo knows that Bo didn't leave.

Though accounts vary with respect to whether neg-raising inferences are explained as a syntactic or a pragmatic phenomenon, all associate these inferences with particular predicates in some way or other---e.g. \textit{think, believe, suppose, imagine, want}, and \textit{expect} are often taken to be associated with neg-raising inferences as a matter of knowledge one has about those predicates, while \textit{say, claim, regret}, and \textit{realize} are not \citep{horn-1971, horn-1978}.

One challenge for such approaches is that whether a neg-raising inference is triggered varies with aspects of the context, such as the predicate's subject---e.g. \ref{ex:variability-present} triggers the inference that the speaker thinks Jo didn't leave---and tense---e.g. \ref{ex:variability-past} does not trigger the same inference as \ref{ex:variability-present}.  

\ex. \label{ex:variability-tense}
\a. \label{ex:variability-present} I don't know that Jo left.
\b. \label{ex:variability-past} I didn't know that Jo left.

While some kinds of variability can be captured by standing accounts, other kinds have yet to be discussed at all. For example, beyond a predicate's subject and tense, the syntactic structure of its clausal complement also appears to matter: \ref{ex:variability-think-passive} and \ref{ex:variability-know-passive} can both trigger neg-raising interpretations, while \ref{ex:variability-think-active} and \ref{ex:variability-know-active} cannot.

\ex. \label{ex:variability-think}
\a. \label{ex:variability-think-passive} Jo wasn't thought to be very intelligent.
\b. \label{ex:variability-think-active} Jo didn't think to get groceries.

\ex.
\a. \label{ex:variability-know-passive} Jo wasn't known to be very intelligent.
\b. \label{ex:variability-know-active} Jo didn't know to get groceries.

Should these facts be chalked up to properties of the predicates in question? Or are they general to how these predicates compose with their complements? These questions are currently difficult to answer for two reasons: (i) there are no existing, lexicon-scale datasets that measure neg-raising across a variety of contexts---e.g. manipulating subject, tense and complement type; and (ii) even if there were, no models currently exist for answering these questions given such a dataset.

We fill this lacuna by (i) collecting a large-scale dataset of neg-raising judgments for effectively all English clause-embedding verbs with a variety of both finite and non-finite complement types; and (ii) extending \citeauthor{white-rawlins-2016}' (\citeyear{white-rawlins-2016}) model of s(emantic)-selection, which induces semantic type signatures from syntactic distribution, with a module that associates semantic types with the inferences they trigger. We use this model to jointly induce semantic types and their relationship to neg-raising inferences, showing that the best fitting model attributes some neg-raising inferences to properties of particular predicates and others to general properties of syntactic structures.\footnote{Data are available at \href{http://megaattitude.io}{megaattitude.io}.}

We begin with background on theoretical approaches to neg-raising, contrasting the two main types of accounts: syntactic and pragmatic (\S\ref{sec:background}). We then present our methodology for measuring neg-raising across a variety of predicates and syntactic contexts (\S\ref{sec:data}) as well as our extension of \citeauthor{white-rawlins-2016}' s-selection model (\S\ref{sec:model}). Finally, we discuss the results of fitting (\S\ref{sec:experiment}) our model to our neg-raising dataset (\S\ref{sec:results}).

\section{Background}
\label{sec:background}

Two main types of approaches have been proposed to account for neg-raising interpretations: syntactic and pragmatic (see \citealt{zeijlstra-2018, crowley-2019} for reviews). We do not attempt to adjudicate between the two here---rather aiming to establish the explanatory devices available to each for later interpretation relative to our modeling results.

\paragraph{Syntactic Approach}

In syntactic approaches, neg-raising interpretations arise from some syntactic relation between a matrix negation and an unpronounced embedded negation that is licensed by the neg-raising predicate. This is classically explained via a syntactic rule that ``raises'' the negation from the subordinate clause to the main clause, as in \ref{ex:negation-raised}, though accounts using alternative syntactic relations exist (\citealt{fillmore-1963, kiparsky-1970, jackendoff-1971, pollack-1976, collins-postal-2014, collins-postal-2017, collins-postal-2018}; cf. \citealp{klima-1964, zeijlstra-2018}; see also \citealp{lasnik-1972}).

\ex.\label{ex:negation-raised}
\begin{dependency}[edge slant=0,baseline]
\begin{deptext}
Jo \& does \& not \& believe \& Bo \& did \& \underline{\phantom{not}} \& leave. \\
\end{deptext}
\depedge[edge unit distance=0.3ex,hide label,edge below,thick]{7}{3}{neg-raising}
\end{dependency}

Evidence for syntactic accounts comes from the distribution of negative polarity items, Horn-clauses, and island phenomena (\citealp{horn-1971, collins-postal-2014, collins-postal-2017, collins-postal-2018}; cf. \citealp{zwarts-1998, gajewski-2011, chierchia-2013, horn-2014, romoli-mandelkern-2019}). 

Purely syntactic approaches to neg-raising have effectively one method for explaining variability in neg-raising inferences relative to subject, tense, and subordinate clause structure (as discussed in \S\ref{sec:introduction}): if a certain lexical item---e.g. \textit{know}---occurs in some sentence that licenses a neg-raising inference---e.g. \ref{ex:variability-know-passive}---and another that doesn't---e.g. \ref{ex:variability-know-active}---one must say that the structure in the first differs from the second in such a way that the first allows the relevant syntactic relation while the second does not. This implies that, even in cases like \ref{ex:variability-present} v. \ref{ex:variability-past}, where there is no apparent structural difference (beyond the subject), the structures differ on some neg-raising-relevant property. This can be implemented by saying that, e.g. the same verb can select for two different structural properties---one that licenses neg-raising and one that does not---or that the verb is somehow ambiguous and its variants differ with respect to some neg-raising-relevant, syntactic property.

\paragraph{Semantic/Pragmatic Approach}

In semantic/pragmatic approaches, neg-raising interpretations are derived from an \textit{excluded middle} (EM or \textit{opinionatedness}) inference \citep{bartsch-1973,horn-1978,horn-bayer-1984,tovena-2001,gajewski-2007, romoli-2013,xiang-2013,homer-2015}. This approach posits that, anytime a neg-raising predicate $v$ is used to relate entity $x$ with proposition $p$, the hearer assumes that either \textit{x v p} or \textit{x v $\lnot$p}. For example, in the case of \textit{believe}, as in \ref{ex:emp-pos}, the hearer would assume that Jo either believes that Bo left or that Bo didn't leave. 

\ex. \label{ex:emp-pos} Jo believes that Bo left.
\a. \textit{truth conditions:} $x$ \textsc{believe} $p$
\b. \textit{inference:}  $x$ \textsc{believe} $p$ $\lor$ $x$ \textsc{believe} $\lnot p$

The EM inference is impotent in the positive cases but drives further inferences in the negative, where the first EM disjunct is cancelled by the truth conditions: if Jo doesn't believe that Bo left and Jo believes that Bo left or that Bo didn't leave, then Jo must believe that Bo didn't leave.  

\ex. \label{ex:emp-neg} Jo doesn't believe that Bo left.
\a. \textit{truth conditions:} $x$ $\lnot$ \textsc{believe} $p$
\b. \textit{inference:}  \sout{$x$ \textsc{believe} $p$ $\lor$} $x$ \textsc{believe} $\lnot p$

To capture non-neg-raising predicates, one must then say that some predicates trigger the EM inference, while others don't \citep{horn-1989}. However, such lexical restrictions alone cannot exhaustively explain the variability in whether verbs trigger presuppositions with certain subjects, as noted for \ref{ex:non-neg-raising} and \ref{ex:variability-present}. To explain this, \citet{gajewski-2007} posits that neg-raising predicates are soft presupposition triggers. Effectively, the EM inferences are defeasible, and when they are cancelled, the neg-raising inference does not go through \citep{abusch-2002}. This is supported by cases of explicit cancellation of the EM inference---e.g. the neg-raising inference \ref{ex:cancel-neg-raising-c} that would otherwise be triggered by \ref{ex:cancel-neg-raising-2} does not go through in the context of \ref{ex:cancel-neg-raising-1}.

\ex. \label{ex:cancel-neg-raising}
\a. \label{ex:cancel-neg-raising-1} Bill doesn't know who killed Caesar. He isn't even sure whether or not Brutus and Caesar lived at the same time. So...
\b. \label{ex:cancel-neg-raising-2} Bill doesn't believe Brutus killed Caesar. \vspace{-2mm}\newline
			\noindent\rule{6cm}{0.4pt}
\c. \label{ex:cancel-neg-raising-c} $\centernot\rightsquigarrow$ Bill believes Brutus didn't kill Caesar.

This sort of explanation relies heavily on semantic properties of particular verbs and naturally covers variability that correlates with subject and tense differences---e.g. \ref{ex:variability-present} v. \ref{ex:variability-past}---since facts about how one discusses their own belief or desire states, in contrast to others belief states, at different times plausibly matter to whether a hearer would make the EM inference. The explanation for variation relative to subordinate clause structure is less clear but roughly two routes are possible: (i) some property of the subordinate clause licenses (or blocks) EM inferences; and/or (ii) predicate ambiguity correlates with which subordinate clause structure (or property thereof) a predicate selects.  

\paragraph{Abstracting the Approaches} Across both approaches, there are roughly three kinds of explanations for neg-raising inferences that can be mixed-and-matched: (i) lexical properties might directly or indirectly (e.g. via an EM inference) license a neg-raising inference; (ii) properties of a subordinate clause structure might directly or indirectly license a neg-raising inference; and/or (iii) lexical and structural properties might interact---e.g. via selection---to directly or indirectly license a neg-raising inference. We incorporate these three kinds of explanation into our models (\S\ref{sec:model}), which we fit to the data described in the next section.

\section{Data}
\label{sec:data}

We develop a method for measuring neg-raising analogous to \citeauthor{white-rawlins-2018}-\citeauthor{white-et-al-2018}'s (\citeyear{white-rawlins-2018}) method for measuring veridicality inferences. With the aim of capturing the range of variability in neg-raising inferences across the lexicon, we deploy this method to test effectively all English clause-embedding verbs in a variety of subordinate clause types---finite and nonfinite---as well as matrix tenses---\textit{past} and \textit{present}---and matrix subjects---\textit{first} and \textit{third person}.


\paragraph{Method} Participants are asked to answer questions like \ref{ex:negraising-question} using a 0-1 slider, wherein the first italicized sentence has negation in the matrix clause and the second italicized sentence has negation in the subordinate.\footnote{The full task instructions are given in Appendix \ref{sec:instructions}.}

\ex. If I were to say \textit{I don't think that a particular thing happened}, how likely is it that I mean \textit{I think that that thing didn't happen}? \label{ex:negraising-question}

Because some sentences, such the italicized in \ref{ex:acceptability-question}, sound odd with negation in the matrix clause,  participants are asked to answer how easy it is to imagine someone actually saying the sentence---again, on a 0-1 slider. The idea here is that the harder it is for participants to imagine hearing a sentence, the less certain they probably are about the judgment to questions like \ref{ex:negraising-question}. 

\ex. How easy is it for you to imagine someone saying \textit{I don't announce that a particular thing happened}? \label{ex:acceptability-question}

Acknowledging the abuse of terminology, we refer to responses to \ref{ex:acceptability-question} as \textit{acceptability responses}. We incorporate these responses into our model (\S\ref{sec:model}) as weights determining how much to pay attention to the corresponding \textit{neg-raising response}.

\paragraph{Materials}

We use the MegaAcceptability dataset of \citet{white-rawlins-2016} as a basis on which to construct acceptable items for our experiment. MegaAcceptability contains ordinal acceptability judgments for 50,000 sentences, including 1,000 clause-embedding English verbs in 50 different syntactic frames. To avoid typicality effects, these frames are constructed to contain as little lexical content as possible besides the verb at hand---a method we follow here. This is done by ensuring that all NP arguments are indefinite pronouns \textit{someone} or \textit{something} and all verbs besides the one being tested are \textit{do}, \textit{have} or \textit{happen}. We focus on the six frames in \ref{ex:thatSactive}--\ref{ex:toVPstativepassive}.

\ex. {\small [NP \underline{\phantom{\_\_}} that S]}\\Someone \underline{knew} that something happened. \label{ex:thatSactive}\vspace{-1mm}

\ex. {\small [NP \underline{\phantom{\_\_}} to VP[+\textsc{ev}]]}\\Someone \underline{liked} to do something. \label{ex:toVPeventiveactive}\vspace{-1mm}

\ex. {\small [NP \underline{\phantom{\_\_}} to VP[-\textsc{ev}]]}\\Someone \underline{wanted} to have something. \label{ex:toVPstativeactive}\vspace{-1mm}

\ex. {\small [NP be \underline{\phantom{\_\_}} that S]}\\Someone was \underline{told} that something happened.\label{ex:thatSpassive}\vspace{-1mm}

\ex. {\small [NP be \underline{\phantom{\_\_}} to VP[+\textsc{ev}]]}\\Someone was \underline{ordered} to do something.\label{ex:toVPeventivepassive}\vspace{-1mm}

\ex. {\small [NP be \underline{\phantom{\_\_}} to VP[-\textsc{ev}]]}\\Someone was \underline{believed} to have something.\label{ex:toVPstativepassive}

These frames were chosen so as to manipulate (i) the presence and absence of tense in the subordinate clause; (ii) the presence or absence of a direct object; and (iii) the lexical aspect of the complement. The frames with direct objects were presented in passivized form so that they were acceptable with both communicative predicates---e.g. \textit{tell}---and emotive predicates---e.g. \textit{sadden}---the latter of which tend to occur with expletive subjects. Lexical aspect was manipulated because some verbs---e.g. \textit{believe}---are more acceptable with nonfinite subordinate clauses headed by a stative than ones headed by an eventive, while others---e.g. \textit{order}---show the opposite pattern.

In light of the variability in neg-raising inferences across the same verb in different tenses---compare again \ref{ex:variability-present} and \ref{ex:variability-past}---we aim to manipulate the matrix tense of each clause-taking verb in our experiment. This is problematic, because the MegaAcceptability dataset only contains items in the past tense. We could simply manipulate the tense for any acceptable sentences based on such past tense items, but some verbs do not sound natural in the present tense with some subordinate clauses---compare the sentences in \ref{ex:pastprescontrast}. 

\ex. \label{ex:pastprescontrast}
\a. Jo wasn't told that Mary left.
\b. Jo isn't told that Mary left.

To remedy this, we extend MegaAcceptability with tense/aspect information by collecting acceptability judgments for modified versions of each sentence in MegaAcceptability, where the target verb is placed in either present or past progressive.\footnote{See Appendix \ref{sec:megaacceptability2} for details.} Combined with MegaAcceptability, our extended dataset results in a total of 75,000 verb-tense-frame pairs: 50,000 from the MegaAcceptability dataset and 25,000 from our dataset. From this combined dataset, we take past and present tense items rated on average 4 out of 7 or better (after rating normalization), for our experiment. This yields 3,968 verb-tense-frame pairs and 925 unique verbs. With our subject manipulation (first v. third person), the number of items doubles, producing 7,936 items. Table \ref{tab:sample-size} summarizes the distribution of verbs in each frame and tense.

To construct items, we follow the method of \citet{white-et-al-2018} of ``bleaching'' all lexical category words in our sentences (besides the subordinate clause-taking verb) by realizing NPs as \textit{a particular person} or \textit{a particular thing}. Verbs are replaced with \textit{do}, \textit{have}, or \textit{happen}. This method aims to avoid unwanted typicality effects that might be introduced by interactions between our predicates of interest and more contentful items elsewhere in the sentence.\footnote{Because this method has not been previously validated for measuring neg-raising, we report two validation experiments in Appendix \ref{sec:validation}, which demonstrate that the measure accords with judgments from prior work.} 

\begin{table}[t!]
\centering
\begin{tabular}{llr}  
\toprule
\textbf{Matrix tense}    & \textbf{Frame} & \textbf{\# verbs} \\
\midrule
{} & {NP \_ that S} & 556 \\
{} & {NP \_ to VP[+\textsc{ev}]} & 400\\ 
{\it past} & {NP \_ to VP[-\textsc{ev}]} & 359\\
{} & {NP be \_ that S} & 255 \\
{} & {NP be \_ to VP[+\textsc{ev}]} & 461\\ 
{} & {NP be \_ to VP[-\textsc{ev}]} & 460\\
\hline
{} & {NP \_ that S} & 413 \\
{} & {NP \_ to VP[+\textsc{ev}]} & 219\\ 
{\it present} & {NP \_ to VP[-\textsc{ev}]} & 155\\
{} & {NP be \_ that S} & 176 \\
{} & {NP be \_ to VP[+\textsc{ev}]} & 268\\ 
{} & {NP be \_ to VP[-\textsc{ev}]} & 246\\
\bottomrule
\end{tabular}
\vspace{-2mm}
\caption{\label{tab:sample-size} \# of verbs acceptable in each tense-frame pair based on our extension of MegaAcceptability.}
\vspace{-5mm}
\end{table}

\begin{figure*}[t!]
    \centering
    \includegraphics[width=1.95\columnwidth]{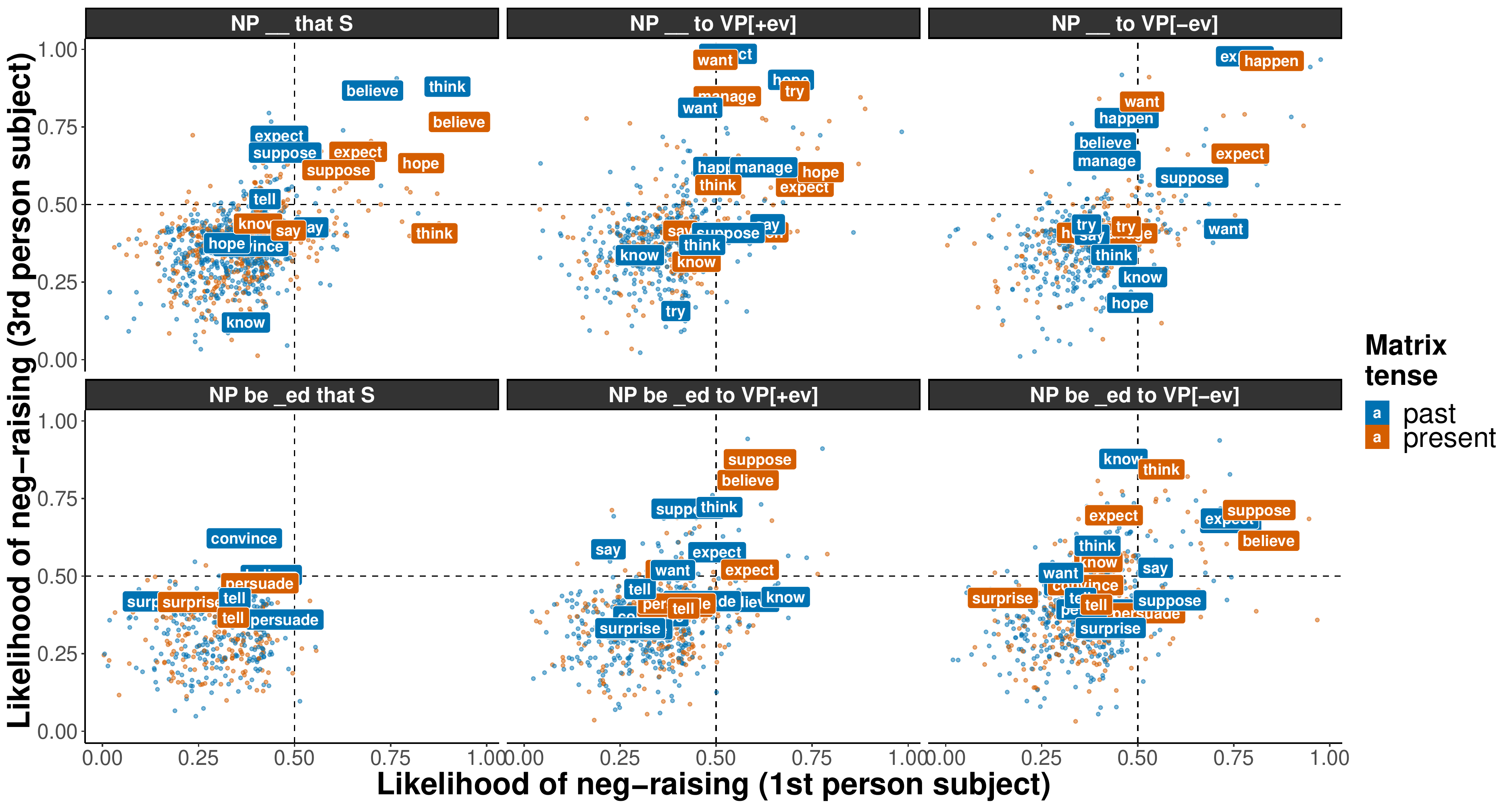}
    \vspace{-2mm}
    \caption{\label{fig:neg-raising-result}Normalized neg-raising scores for different subject, tense, and frame pairs.}
    \vspace{-6mm}
\end{figure*}

We partition items into 248 lists of 32 items. Each list is constrained such that (i) 16 items had a first person subject, and 16 items had a third person subject; (ii) 16 items contain a \textit{low frequency} verb and 16 items contain a \textit{high frequency} verb, based on a median split of the frequencies in the \textsc{subtlex\_us} word frequency database \citep{subtlex-us-2009}; (iii) 16 items are \textit{low acceptability} and 16 items are \textit{high acceptability}, based on a median split of the normalized acceptabilities for items selected from our extension of the MegaAcceptability dataset; (iv) no verb occurred more than once in the same list; (v) items containing a particular combination of matrix tense and syntactic frame occur in rough proportion to the number of verbs that are acceptable with that tense-frame combination based on our extension of the MegaAcceptability dataset (Table \ref{tab:sample-size}).

\paragraph{Participants}
1,108 participants were recruited through Amazon Mechanical Turk to give 10 ratings per sentence in the 248 lists of 32---i.e. the end result contains 79,360 ratings for each of neg-raising and acceptability judgments. Participants were not allowed to respond to the same list more than once, though they were allowed to respond to as many lists as they liked. Each participants responded to 2.3 lists on average (min: 1, max: 16, median: 1). Of the 1,108 participants, 10 reported not speaking American English as their native language. Responses from these participants were filtered from the dataset prior to analysis. From this, responses for 27 lists were lost ($\sim$1\% of the responses). This filtering removed at most two judgments for any particular item. 

\paragraph{Results}
Figure \ref{fig:neg-raising-result} plots the normalized neg-raising scores for verbs in different subject (axes)-tense (color)-frame (block) contexts.\footnote{See Appendix \ref{sec:normalization} for details on normalization.} A verb (in some tense) being toward the top-right corner means that it shows strong neg-raising inferences with both first person and third person subjects, while a verb being towards the bottom-right corner means that it shows neg-raising behavior with first person subjects but not with third person subjects. The converse holds for the top-left corner: neg-raising behavior is seen with third person subjects but not first. We see that our method correctly captures canonical neg-raising predicates---e.g. \textit{think} and \textit{believe} with finite complements and \textit{want} and \textit{expect} with infinitival complements---as well as canonical non-neg-raising predicates---e.g. \textit{know} and \textit{say} with finite complements and \textit{try} and \textit{manage} with infinitivals.

\section{Model}
\label{sec:model}
We aim to use our neg-raising dataset to assess which aspects of neg-raising inferences are due to properties of lexical items and which aspects are due to properties of the structures they compose with. To do this, we extend \citeauthor{white-rawlins-2016}' (\citeyear{white-rawlins-2016}) model of s(emantic)-selection, which induces semantic type signatures from syntactic distribution, with a module that associates semantic types with the \textit{inference patterns} they trigger. 

Our model has two hyperparameters that correspond to the theoretical constructs of interest: (i) the number of lexical properties relevant to neg-raising; and (ii) the number of structural properties relevant to neg-raising. In \S\ref{sec:experiment}, we report on experiments aimed at finding the optimal setting of these two hyperparameters, and we analyze the parameters of the model fit corresponding to these hyperparameters in \S\ref{sec:results}. 

\paragraph{S-selection Model}
\citeauthor{white-rawlins-2016}' (\citeyear{white-rawlins-2016}) model of s-selection aims to induce verbs' semantic type signatures---e.g. that \textit{love} can denote a relation between two entities and \textit{think} can denote a relation between an entity and a proposition---from their syntactic distribution---e.g. that \textit{love} is acceptable in NP \underline{\phantom{\_\_}} NP frames and that \textit{think} is acceptable in NP \underline{\phantom{\_\_}} S frames. They formalize this task as a boolean matrix factorization (BMF) problem: given a boolean matrix $\mathbf{D} \in \mathbb{B}^{|\mathcal{V}| \times |\mathcal{F}|} = \{0, 1\}^{|\mathcal{V}| \times |\mathcal{F}|}$, wherein $d_{vf} = 1$ iff verb $v \in \mathcal{V}$ is acceptable in syntactic frame $f \in \mathcal{F}$, one must induce boolean matrices $\bm\Lambda \in \mathbb{B}^{|\mathcal{V}| \times |\mathcal{T}|}$ and $\bm\Pi \in \mathbb{B}^{|\mathcal{T}| \times |\mathcal{F}|}$, wherein $\lambda_{vt} = 1$ iff verb $v$ can have semantic type signature $t \in \mathcal{T}$ and $\pi_{tf} = 1$ iff $t$ can be mapped onto syntactic frame $f$, such that \ref{ex:whiteetalequation}: verb $v$ is acceptable in frame $f$ iff $v$ has some type $t$ that can be mapped (or \textit{projected}) onto $f$.

\ex. $d_{vf} \approx \bigvee_t \lambda_{vt} \land \pi_{tf}$ \label{ex:whiteetalequation}

As is standard in matrix factorization, the equivalence is approximate and is only guaranteed when there are as many semantic type signatures $\mathcal{T}$ as there are frames $\mathcal{F}$, in which case, the best solution is the one with $\bm\Lambda = \mathbf{D}$ and $\bm\Pi$ as the identity matrix of dimension $|\mathcal{T}| = |\mathcal{F}|$. Because this solution is trivial, $|\mathcal{T}|$ is generally much smaller than $|\mathcal{F}|$ and determined by fit to the data---in BMF, the count of how often $d_{vf} \neq \bigvee_t \lambda_{vt} \land \pi_{tf}$.

As an estimate of $\mathbf{D}$, \citeauthor{white-rawlins-2016} use the MegaAcceptability dataset, which we use in constructing our neg-raising dataset (\S\ref{sec:data}). Instead of directly estimating the boolean matrices $\bm\Lambda$ and $\bm\Pi$, they estimate a probability distribution over the two under the strong independence assumption that all values $\lambda_{vt}$ and $\pi_{tf}$ are pairwise independent of all other values. This implies \ref{ex:whiteetalprobability}.\footnote{See Appendix \ref{sec:modelappendix} for the derivation of \ref{ex:whiteetalprobability}.}

\ex. $\mathbb{P}(d_{vf}) = 1-\prod_t 1-\mathbb{P}(\lambda_{vt})\mathbb{P}(\pi_{tf})$ \label{ex:whiteetalprobability}

\citeauthor{white-rawlins-2016} treat $\mathbb{P}(d_{vf})$ as a fixed effect in an ordinal mixed effects model, which provides the loss function against which $\mathbb{P}(\lambda_{vt})$ and $\mathbb{P}(\pi_{tf})$ are optimized. They select the number of semantic type signatures to analyze by setting $|\mathcal{T}|$ such that an information criterion is optimized. 

\paragraph{Neg-Raising Model}
We retain the main components of \citeauthor{white-rawlins-2016}' model but add a notion of \textit{inference patterns} associated both with properties of verbs, on the one hand, and with semantic type signatures, on the other. In effect, this addition models inferences, such as neg-raising, as arising via a confluence of three factors: (i) properties of the relation a lexical item denotes---e.g. in a semantic/pragmatic approach, whatever property of a predicate triggers EM inferences; (ii) properties of the kinds of things that a predicate (or its denotation) relates---e.g. in a syntactic approach, whatever licenses ``raising'' of the negation; and (iii) whether a particular verb has a particular type signature. With respect to (ii) and (iii), it is important to note at the outset that, because we do not attempt to model acceptability, semantic type signatures play a somewhat different role in our model than in \citeauthor{white-rawlins-2016}': instead of determining which structures a verb is compatible with---i.e. (non)finite subordinate clauses, presence of a direct object, etc.---our model's type signatures control the inferences a particular verb can trigger when taking a particular structure. As such, our model's semantic type signatures might be more easily construed as properties of a structure that may or may not license neg-raising.\footnote{Alternatively, they might be construed as (potentially cross-cutting) classes of syntactic structures and/or semantic type signatures that could be further refined by jointly modeling acceptability (e.g. as measured by MegaAcceptability) alongside our measure of neg-raising inferences.} We thus refer to them as \textit{structural properties}---in contrast to predicates' \textit{lexical properties}.

Our extension requires the addition of three formal components to \citeauthor{white-rawlins-2016}' model: (i) a boolean matrix $\bm{\Psi} \in \mathbb{B}^{|\mathcal{V}| \times |\mathcal{I}|}$, wherein $\psi_{vi} = 1$ iff verb $v \in \mathcal{V}$ has property $i \in \mathcal{I}$; (ii) a boolean tensor $\bm{\Phi} \in \mathbb{B}^{|\mathcal{I}| \times |\mathcal{J}| \times |\mathcal{K}|}$, wherein $\phi_{ijk} = 1$ iff property $i$ licenses a neg-raising inference with subject $j \in \mathcal{J}$ and tense $k \in \mathcal{K}$; and (iii) a boolean tensor $\bm{\Omega} \in \mathbb{B}^{|\mathcal{T}| \times |\mathcal{J}| \times |\mathcal{K}|}$, wherein $\omega_{tjk} = 1$ iff semantic type signature $t \in \mathcal{T}$ licenses a neg-raising inference with subject $j$ and tense $k$. 

As it stands, this formulation presupposes that there are both lexical and structural properties relevant to neg-raising. To capture the possibility that there may be only one or the other relevant to neg-raising, we additionally consider two families of \textit{boundary models}. In the boundary models that posit no lexical properties---which (abusing notation) we refer to as $|\mathcal{I}| = 0$---we fix $\bm\Psi = \mathbf{1}_{|\mathcal{V}|}$ and $\bm\Phi = \mathbf{1}_{|\mathcal{I}|} \otimes \mathbf{1}_{|\mathcal{J}|} \otimes \mathbf{1}_{|\mathcal{K}|}$. In the boundary models that posit no structural properties ($|\mathcal{T}| = 0$) we fix $\bm\Pi = \mathbf{1}_{|\mathcal{F}|}$, $\bm\Lambda = \mathbf{1}_{|\mathcal{V}|}$, and $\bm\Omega = \mathbf{1}_{|\mathcal{T}|} \otimes \mathbf{1}_{|\mathcal{J}|} \otimes \mathbf{1}_{|\mathcal{K}|}$. 

Analogous to \citeauthor{white-rawlins-2016}, we treat our task as a problem of finding $\bm\Lambda, \bm\Pi, \bm{\Psi}, \bm{\Phi}, \bm{\Omega}$ that best approximate the tensor $\mathbf{N}$, wherein $n_{vfjk} = 1$ iff verb $v$ licenses neg-raising inferences in frame $f$ with subject $j$ and tense $k$. This is formalized in \ref{ex:ourbooleanequation}, which implies that $n_{vfjk} = 1$ iff there is some pairing of semantic type signature $t$ and inference pattern $i$ such that (i) verb $v$ has semantic type signature $t$; (ii) verb $v$ licenses inference pattern $i$; (iii) semantic type signature $t$ can map onto frame $f$; and (iv) both $t$ and $i$ license a neg-raising inference with subject $j$ and tense $k$.

\ex. $n_{vfjk} \approx \bigvee_{t,i} \lambda_{vt} \land \psi_{vi} \land \phi_{ijk} \land \pi_{tf} \land \omega_{tjk}$ \label{ex:ourbooleanequation}

Also analogous to \citeauthor{white-rawlins-2016}, we aim to estimate $\mathbb{P}(n_{vfjk})$ (rather than $n_{vfjk}$ directly) under similarly strong independence assumptions: $\mathbb{P}(\lambda_{vt},\psi_{vi},\phi_{ijk}, \pi_{tf}, \omega_{tjk}) = \mathbb{P}(\lambda_{vt})\mathbb{P}(\psi_{vi})$ $\mathbb{P}(\phi_{ijk})\mathbb{P}(\pi_{tf})\mathbb{P}(\omega_{tjk}) = \zeta_{vtifjk}$, implying \ref{ex:ourprobabilisticequation}.

\ex. $\mathbb{P}(n_{vfjk}) = 1 - \prod_{t,i} 1 - \zeta_{vtifjk}$ \label{ex:ourprobabilisticequation}

We design the loss function against which $\mathbb{P}(\lambda_{vt})$, $\mathbb{P}(\psi_{vi})$, $\mathbb{P}(\phi_{ijk})$, $\mathbb{P}(\pi_{tf})$, and $\mathbb{P}(\omega_{tjk})$ are optimized such that (i) $\mathbb{P}(n_{vfjk})$ is monotonically related to the neg-raising response $r_{vfjkl}$ given by participant $l$ for an item containing verb $v$ in frame $f$ with subject $j$ and tense $k$ (if one exists); but (ii) participants may have different ways of using the response scale. For example, some participants may prefer to use only values close to 0 or 1, while others may prefer values near 0.5; or some participants may be prefer lower likelihood values while others may prefer higher values. To implement this, we incorporate (i) a fixed scaling term $\sigma_0$; (ii) a fixed shifting term $\beta_0$; (iii) a random scaling term $\sigma_l$ for each participant $l$; and (iv) a random shifting term $\beta_l$ for each participant $l$. We define the expectation for a response $r_{vfjkl}$ as in \ref{ex:nrev}.

\ex. $\hat{r}_{vfjkl} = \mathrm{logit}^{-1}\left(m_a\nu_{vfjk} + \beta_0 + \beta_l\right)$\\where $\nu_{vfjk} = \mathrm{logit}\left(\mathbb{P}(n_{vfjk})\right)$\\
\phantom{where} $m_l = \exp\left(\sigma_0+\sigma_l\right)$\label{ex:nrev}

We optimize $\mathbb{P}(\lambda_{vt})$, $\mathbb{P}(\psi_{vi})$, $\mathbb{P}(\phi_{ijk})$, $\mathbb{P}(\pi_{tf})$, and $\mathbb{P}(\omega_{tjk})$ against a KL divergence loss, wherein $r_{vfjkl}$ is taken to parameterize the true distribution and $\hat{r}_{vfjkl}$ the approximating distribution. 

\ex. $\mathrm{D}(r\;\|\;\hat{r}) = -\left[r\log\frac{\hat{r}}{r} + (1-r)\log\frac{1-\hat{r}}{1-r}\right]$

\vspace{-1mm}

\noindent To take into account that it is harder to judge the neg-raising inferences for items that one cannot imagine hearing used, we additionally weight the above-mentioned KL loss by a normalization of the acceptability responses for an item containing verb $v$ in frame $f$ with subject $j$ and tense $k$. We infer this value from the acceptability responses for an item containing verb $v$ in frame $f$ with subject $j$ and tense $k$ given by participant $l$, assuming a form for the expected value of $a_{vfjkl}$ as in \ref{ex:accev}---analogous to \ref{ex:nrev}. (Unlike $\nu_{vfjk}$ in \ref{ex:nrev}, $\alpha_{vfjk}$ in \ref{ex:accev} is directly optimized.)

\ex. $\hat{a}_{vfjkl} = \mathrm{logit}^{-1}\left(m'_l\alpha_{vfjk} + \beta'_0 + \beta'_l\right)$\\where $m'_l = \exp\left(\sigma'_0+\sigma'_l\right)$\label{ex:accev}

The final loss against which $\mathbb{P}(\lambda_{vt})$, $\mathbb{P}(\psi_{vi})$, $\mathbb{P}(\phi_{ijk})$, $\mathbb{P}(\pi_{tf})$, $\mathbb{P}(\omega_{tjk})$ are optimized is \ref{ex:finalloss}.\footnote{An additional term (not shown) is added to encode the standard assumption that the random effects terms are normally distributed with mean 0 and unknown variance.}

\ex. \label{ex:finalloss} $\mathcal{L} = \sum \alpha'_{vfjk}\mathrm{D}(r_{vfjkl}\;\|\;\hat{r}_{vfjkl})$

where $\alpha'_{vfjk} = \mathrm{logit}^{-1}(\alpha_{vfjk})$.

\section{Experiment}
\label{sec:experiment}

\begin{figure}[t]
    \centering
    \includegraphics[width=\columnwidth]{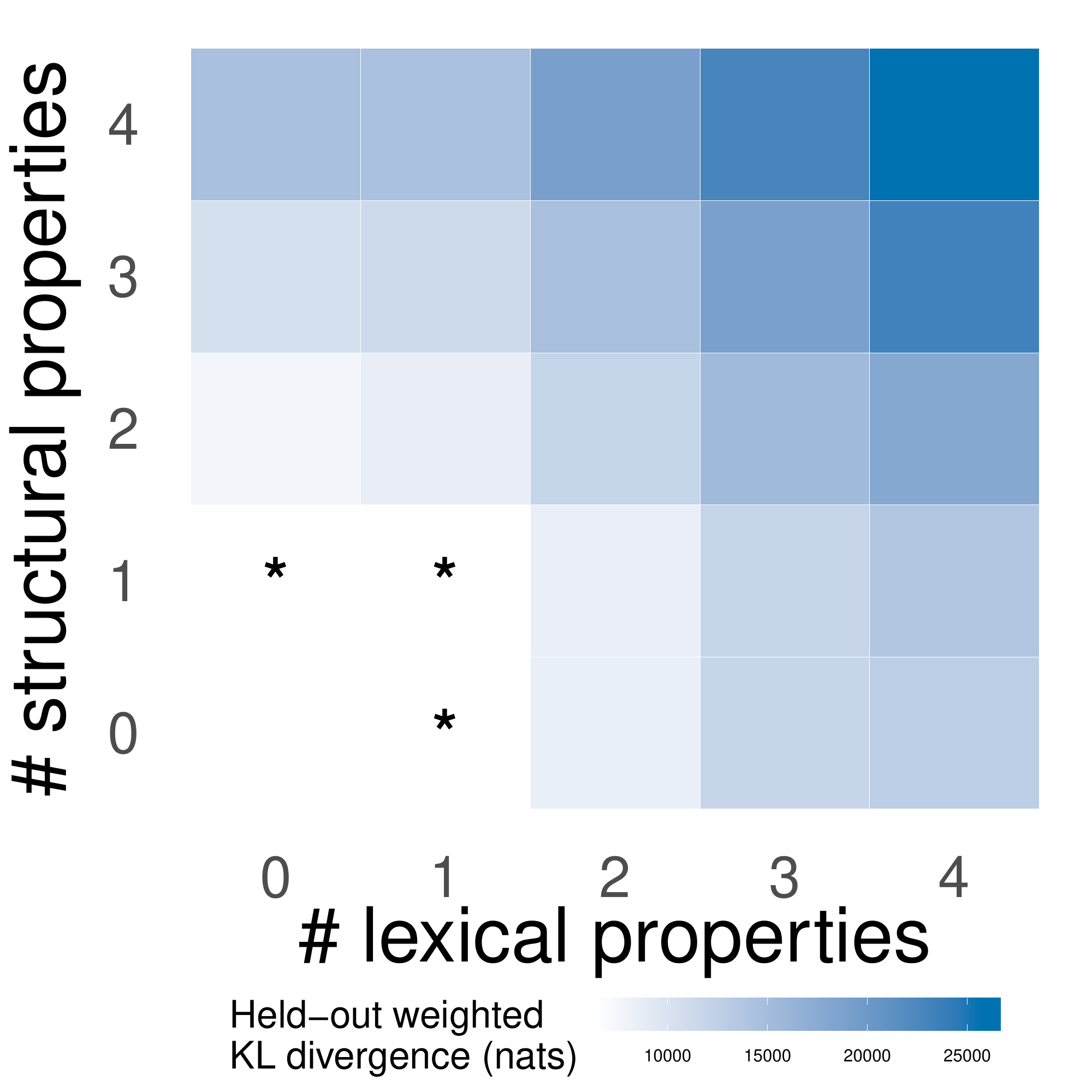}
    \vspace{-8mm}
    \caption{Sum of the weighted KL divergence loss across all five folds of the cross-validation for each setting of $|\mathcal{I}|$ (\# of lexical properties) and $|\mathcal{T}|$ (\# of structural properties). $|\mathcal{I}| = |\mathcal{T}| = 0$ was not run.}
    \label{fig:crossvalidation}
    \vspace{-5mm}
\end{figure}

We aim to find the optimal settings, relative to our neg-raising data, for (i) the number $|\mathcal{I}|$ of lexical properties relevant to neg-raising that it assumes; and (ii) the number $|\mathcal{T}|$ of structural properties relevant to neg-raising that it assumes. As with other models based on matrix factorization, higher values for $|\mathcal{I}|$ (with a fixed $|\mathcal{T}|$) or $|\mathcal{T}|$ (with a fixed $|\mathcal{I}|$) will necessarily fit the data as well or better than lower values, since a model with larger $|\mathcal{I}|$ or $|\mathcal{T}|$ can embed the model with a smaller value. However, this better fit comes at the cost of increased risk of overfitting due to the inclusion of superfluous dimensions. To mitigate the effects of overfitting, we conduct a five-fold cross-validation and select the model(s) with the best performance (in terms of our weighted loss) on held-out data.

\paragraph{Method}

In this cross-validation, we pseudorandomly partition sentences from the neg-raising experiments into five sets (folds), fit the model with some setting of $|\mathcal{I}|, |\mathcal{T}|$ to the neg-raising responses for sentences in four of these sets (80\% of the data), then compute the loss on the held-out set---repeating with each partition acting as the held-out set once. The assignment of items to folds is pseudorandom in that each fold is constrained to contain at least one instance of a particular verb with a particular complement type in some tense with some subject. If such a constraint were not enforced, on some folds, the model would have no data upon which to predict that verb with that complement. We consider each possible pairing of $|\mathcal{I}|, |\mathcal{T}| \in \{0, 1, 2, 3, 4\}$, except $|\mathcal{I}| = |\mathcal{T}| = 0$. The same partitioning is used for every setting of $|\mathcal{I}|$ and $|\mathcal{T}|$, enabling paired comparison by sentence. 

\paragraph{Implementation}

We implement our model in \texttt{tensorflow 1.14.0} \citep{abadi-2016}. We use the Adam optimizer \citep{kingma-ba-2015} with a learning rate of 0.01 and default hyperparameters otherwise.

\paragraph{Results}

Figure \ref{fig:crossvalidation} plots the sum of the weighted KL divergence loss across all five folds of the cross-validation for each setting of $|\mathcal{I}|$ (number of lexical properties) and $|\mathcal{T}|$ (number of structural properties). The best-performing models in terms of held-out loss (starred in Figure \ref{fig:crossvalidation}) are (in order): (i) one that posits one lexical property and no structural properties; (ii) one that posits no lexical properties and one structural property; and (iii) one that posits one lexical property and one structural property. None of these models' performance is reliably different from the others---as determined by a nonparametric bootstrap computing the 95\% confidence interval for the pairwise difference in held-out loss between each pairing among the three---but all three perform reliably better than all other models tested. 

Among these three, the model with the best fit to the dataset has $|\mathcal{I}| = 1$ and $|\mathcal{T}| = 1$. This result suggests that neg-raising is not purely a product of lexical knowledge: properties of the subordinate clause that a predicate combines with also influence whether neg-raising inferences are triggered. This is a surprising finding from the perspective of prior work, since (to our knowledge) no existing proposals posit that syntactic properties like the ones we manipulated to build our dataset---i.e. the presence or absence of tense, the presence or absence of an overt subject of the subordinate clause, and eventivity/stativity of a predicate in the subordinate clause---can influence whether neg-raising inferences are triggered. We next turn to analysis of this model fit to understand how our model captures patterns in the data.


\begin{figure*}[t!]
    \centering
    \includegraphics[width=2\columnwidth]{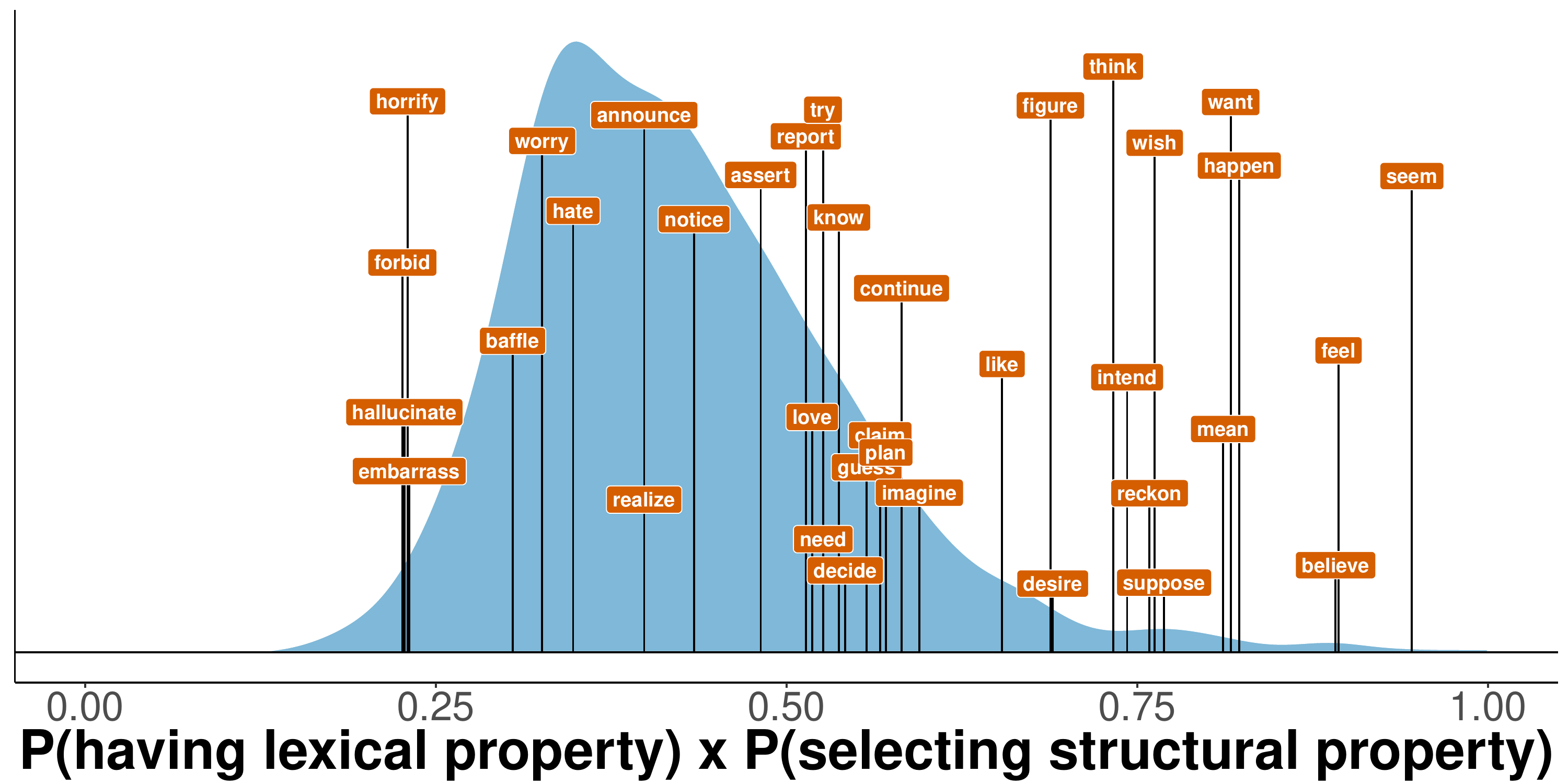}
    \vspace{-2mm}
    \caption{Distribution of $\mathbb{P}(\psi_{vi}) \times \mathbb{P}(\lambda_{vt})$ across predicates, along with selected neg-raising (toward right) and non-neg-raising (toward left) predicates in $|\mathcal{I}| = |\mathcal{T}| = 1$ model.  (Label height is jittered to avoid overplotting.)}
    \label{fig:type-inference}
    \vspace{-5mm}
\end{figure*}

\section{Analysis}
\label{sec:results}

Table \ref{tab:structure-verb-types} gives the $|\mathcal{I}| = |\mathcal{T}| = 1$ model's estimate of the relationship between neg-raising inferences and lexical $\mathbb{P}(\phi_{ijk})$ (top) and structural properties $\mathbb{P}(\omega_{tjk})$ (bottom) with different subjects and tenses. The fact that all of the values in Table \ref{tab:structure-verb-types} are near 1 suggests that predicates having the lexical property or structures having the structural property will give rise to neg-raising inferences regardless of the subject and tense.\footnote{These tables appear to be copies of each other, but they are not. What is happening here is that the model is learning to associate $\mathbb{P}(\phi_{ijk})$ and $\mathbb{P}(\omega_{tjk})$ with (roughly) the square root of the largest expected value across all predicates for the neg-raising response to sentences with subject $j$ and tense $k$. (It sets these values to the square root of the largest expected value because they will be multiplied together.) This strategy allows the model to simply vary $\mathbb{P}(\lambda_{vt})$, $\mathbb{P}(\psi_{vi})$, and $\mathbb{P}(\pi_{tf})$ to capture the likelihood a particular predicate or structure gives rise to neg-raising inferences, as described below.\label{fn:copy}}

This pattern is interesting because it suggests that the model does not capture the variability across different subjects and tenses observed in Figure \ref{fig:neg-raising-result} as a matter of either lexical or structural properties. That is, the model treats any variability in neg-raising inferences across different subjects and/or tenses as an idiosyncratic fact about the lexical item and the structure it occurs with---i.e. noise. This result makes intuitive sense insofar as such variability arises due to pragmatic reasoning that is specific to particular predicates, as opposed to some general semantic property.

\begin{table}[t]
    \centering
\begin{tabular}{llrr}
\toprule
     & & \multicolumn{2}{c}{\textbf{Tense}} \\
\textbf{Property} & \textbf{Person} &  \textit{past} &  \textit{present}         \\
\midrule
\multirow{2}{*}{\it lexical} & \textit{first} &  0.93 &     0.98 \\
     & \textit{third} &  0.95 &     0.98 \\
\midrule
\multirow{2}{*}{\it structural} & \textit{first} &  0.93 &     0.98 \\
     & \textit{third} &  0.95 &     0.98 \\
\bottomrule
\end{tabular}
    \vspace{-2mm}
    \caption{Relationship between neg-raising inferences and lexical property $\mathbb{P}(\phi_{ijk})$ (top) and structural property $\mathbb{P}(\omega_{tjk})$ (bottom) with different subjects and tenses in $|\mathcal{I}| = |\mathcal{T}| = 1$ model.}
    \label{tab:structure-verb-types}
    \vspace{-4mm}
\end{table}



But while the model does not distinguish among neg-raising inference with various subject and tense combinations, it does capture the coarser neg-raising v. non-neg-raising distinction among predicates---namely, by varying the probability that different lexical items have the lexical property $\mathbb{P}(\psi_{vi})$ and the probability that they select the structural property $\mathbb{P}(\lambda_{vt})$. Figure \ref{fig:type-inference} plots the distribution of $\mathbb{P}(\psi_{vi}) \times \mathbb{P}(\lambda_{vt})$ across predicates.\footnote{We plot the distribution of $\mathbb{P}(\psi_{vi}) \times \mathbb{P}(\lambda_{vt})$, instead of showing a scatter plot, because these probabilities show extremely high positive rank correlation---approximately 1. This happens because, when there is only one lexical property and one structural property, the lexical property and selection probabilities are effectively a single parameter $p$, with $\mathbb{P}(\psi_{vi})$ and $\mathbb{P}(\lambda_{vt})$ themselves being set to $\sqrt{p}$ (see also Footnote \ref{fn:copy}).} We see that predicates standardly described as neg-raising (\textit{think}, \textit{believe}, \textit{want}, \textit{seem}, \textit{feel}, etc.) fall to the right, while those standardly described as non-neg-raising (\textit{know}, \textit{notice}, \textit{realize}, \textit{love}, etc.) fall to left. Thus, in some sense, a predicate's probability of having the model's single lexical property (plus its probability of selecting the single structural property) appears to capture something like the probability of neg-raising.   

\begin{table}[h!]
    \centering

\begin{tabular}{lrr}
\begin{tabular}{lr}
\toprule
                \textbf{Structure} &    \textbf{Probability} \\
\midrule
         NP \_\_ that S &  0.91 \\
     NP be \_ed that S &  0.84 \\
     NP \_\_ to VP[+ev] &  0.98 \\
 NP be \_ed to VP[+ev] &  0.93 \\
     NP \_\_ to VP[-ev] &  0.94 \\
 NP be \_ed to VP[-ev] &  0.98 \\
\bottomrule
\end{tabular}
\end{tabular}
\vspace{-3mm}
    \caption{Relationship between structural property and structures $\mathbb{P}(\pi_{tf})$ in $|\mathcal{I}| = |\mathcal{T}| = 1$ model.}
    \label{tab:frame-structure-types}
    \vspace{-2mm}
\end{table}

\noindent The model captures variability with respect to different syntactic structures by modulating $\mathbb{P}(\pi_{tf})$, shown in Table \ref{tab:frame-structure-types}. Looking back to Figure \ref{fig:neg-raising-result}, these values roughly correlate with the largest neg-raising response (across subjects and tenses) seen in that frame, with NP be \_ed that S showing the lowest such value. The value of $\mathbb{P}(\pi_{tf})$ is not the \textit{same} as the largest neg-raising value in Figure \ref{fig:neg-raising-result}, likely due to the fact that many of the predicates that occur in that frame also have small values for $\mathbb{P}(\psi_{vi}) \times \mathbb{P}(\lambda_{vt})$, and thus, when $\mathbb{P}(\pi_{tf})$ is multiplied by that values, it is small. 

\section{Conclusion}
\label{sec:conclusion}
We presented a probabilistic model to induce the mappings from lexical sources and their grammatical sources to neg-raising inferences. We trained this model on a large-scale dataset of neg-raising judgments that we collected for 925 English clause-embedding verbs in six distinct syntactic frames as well as various matrix tenses and subjects. Our model fit the best when positing one lexical property and one structural property. This is a surprising finding from the perspective of prior work, since (to our knowledge) no existing proposals posit that syntactic properties like the ones we manipulated to build our dataset---i.e. the presence or absence of tense, the presence or absence of an overt subject of the subordinate clause, and eventivity/stativity of a predicate in the subordinate clause---can influence whether neg-raising inferences are triggered. Our findings suggest new directions for theoretical research attempting to explain the interaction between lexical and structural factors in neg-raising. Future work in this vein might extend the model proposed here to investigate the relationship between neg-raising and acceptability as well as other related phenomena with associated large-scale datasets, such as lexically triggered veridicality inferences \citep{white-rawlins-2018,white-et-al-2018,white-2019}.

\section*{Acknowledgments}

We would like to thank the FACTS.lab at UR as well as three anonymous reviewers for useful comments.
This work was supported by an NSF grant (BCS-1748969/BCS-1749025) \textit{The
MegaAttitude Project: Investigating selection and polysemy at the scale of the lexicon}.

\bibliography{scil2020}
\bibliographystyle{acl_natbib}


\appendix

\section{Instructions}
\label{sec:instructions}
In this experiment, you will be asked to answer questions about what a person is likely to mean if they say a particular sentence.

Your task will be to respond about the likelihood on the slider that will appear under each question, where the left side corresponds to \textit{extremely unlikely} and the right side corresponds to \textit{extremely likely}.

For instance, you might get the question \textit{If I were to say John has three kids, how likely is it that I mean John has exactly three kids?} with a slider. In this case you would move the slider handle fairly far to the right (toward \textit{extremely likely}), since if someone says "John has three kids", it's pretty likely that they mean that John has exactly three children.

If the question were \textit{If I were to say some of the boys left, how likely is it that I mean all of the boys left?}, then you might move the slider pretty far to the left (toward \textit{extremely unlikely}), since it would be odd if someone says "Some of the boys left – and by that, I mean all of the boys left".

And if the question were \textit{If I were to say Ann didn’t greet everyone politely, how likely is it that I mean Ann was unwelcoming to every single person?}, you might leave the slider in the middle (which corresponds to \textit{maybe or maybe not}), since quite often such sentence can be used to mean Ann greeted some people politely but not all, or to mean Ann was not polite to every single person.

Try to answer the questions as quickly and accurately as possible. Many of the sentences may not be sentences that you can imagine someone ever saying. Try your best to interpret what a speaker would mean in using them. After each question, you will be given a chance to tell us whether the sentence you just responded to isn't something you can imagine a native English speaker ever saying.

Not all questions have correct answers, but a subset in each HIT do. Prior to approval, we check the answers given for this subset. We will reject work containing a substantial number of answers that do not agree with the correct answer.

When the experiment is over, a screen will appear telling you that you are done, and a submission button will be revealed.

\section{Data}
\label{sec:megaacceptability2}
We extend \citeauthor{white-rawlins-2016}' (\citeyear{white-rawlins-2016}) MegaAcceptability v1.0 dataset by collecting acceptability judgments for sentences in present and past progressive tenses---resulting in MegAcceptability v2.0, which subsumes MegaAcceptability v1.0. To enable comparison of the judgments given in MegaAcceptability v1.0 and those we collect, we run an additional \textit{linking experiment} with half items from MegaAcceptability v1.0 and our extension. We then normalize all three datasets separately using the procedure described in \citealt{white-rawlins-2019} and then combine them by using the linking experiment data to train a model to map them into a comparable normalized rating space. Both the extended MegaAcceptability and linking datasets are available at \href{http://megaattitude.io}{megaattitude.io}.

\paragraph{Extended MegaAcceptability}
Our test items are selected and modified from the top 25\% most acceptable verb-frame pairs from the MegaAcceptability dataset of \citet{white-rawlins-2016}, determined by a modified version of the normalization procedure used in \citealt{white-rawlins-2019}. This item set thus contains 12,500 verb-frame pairs, with 1000 unique verbs and the same 50 subcategorization frames (35 in active voice and 15 in passive voice) that are used in MegaAcceptability. 

Given the 12,500 verb-frame pairs, we construct new sentences in both present and past progressive tense/aspect, resulting in a total of 25,000 items. Examples of two sentences from MegaAcceptability v1.0 are given in \ref{ex:acceptability-past} and the corresponding present and past progressive versions are given in \ref{ex:acceptability-present} and \ref{ex:acceptability-pastprogressive}, respectively.

\ex. \label{ex:acceptability-past}
\a. Someone knew which thing to do. 
\b. Someone talked about something. 

\ex. \label{ex:acceptability-present}
\a. Someone knows which thing to do. 
\b. Someone talks about something. 

\ex. \label{ex:acceptability-pastprogressive}
\a. Someone is knowing which thing to do. 
\b. Someone was talking about something.

All methods follow \citealt{white-rawlins-2016}. Sentences are partitioned into 500 lists of 50, with each list constructed such that (i) each frame shows up once in a list, making each list contain 50 unique frames, if possible; (ii) otherwise, the distribution of frames are kept as similar as possible across lists; and (iii) no verbs appear more than once in a list. We gather 5 acceptability judgments per sentence, yielding a total of 125,000 judgments for 25,000 items. 

Judgments for each sentence in a list are collected on a 1-to-7 scale. To avoid typicality effects, we construct the frames to contain as little lexical content as possible besides the verb at hand. For this, we instantiate all NP arguments as indefinite pronouns \textit{someone} or \textit{something} and all verbs besides the one being tested as \textit{do} or \textit{happen}. 565 participants were recruited from Amazon Mechanical Turk, where 562 speak American English as their native language.

\paragraph{Linking experiment}
Because our extension of MegaAcceptability was built in such a way that it likely contains higher acceptability items, the ratings in MegaAcceptability v1.0 and the ratings in our extension are likely not comparable---i.e. a rating in MegaAcceptability v1.0 is, in some sense, a worse rating than in our extension, since our sentences are, by construction, better overall. To put the existing MegaAcceptability dataset and our extended dataset on a comparable scale, we run another experiment to assist in mapping the two datasets to such a comparable scale. We choose 50 items, each with a unique verb, by selecting 26 items from our dataset (14 in present tense and 12 in past progressive tense) and 24 items from MegaAcceptability (all past tense). 

This item selection was constrained such that half of the items chosen were below the median acceptability score and half were above, evenly split across items from our experiment and items from MegaAcceptability v1.0. The items with the lowest acceptability scores consist of 8 in the present, 6 in the past progressive, and 12 in the past tense and so do the items with the highest acceptability scores.  Example items with the low acceptability scores (under this criterion) are shown in \ref{ex:low}, and example items with high acceptability scores are shown in \ref{ex:high}.

\ex. \label{ex:low}
\a. Someone demands about whether something happened. 
\b. Someone was judging to someone that something happened.
\c. Someone invited which thing to do. \label{ex:low-past}

\ex. \label{ex:high}
\a. Someone is distracted. 
\b. Someone was teaching.
\c. Someone dared to do something.

The linking experiment is built in a very similar manner to our extension of MegaAcceptability, described above. Ordinal acceptability judgments are collected on a 1-to-7 scale. 50 participants were recruited to rate all 50 items in the experiment. All of the 50 participants report speaking American English as their native language.

After running the linking experiment, we normalize the ratings in all three datasets separately using a modified version of the procedure described in \citealt{white-rawlins-2019}. Then, we construct one mapping from the normalized ratings in our extension of MegaAcceptability to the normalized ratings for the linking dataset and another mapping from the normalized ratings in the linking dataset to the normalized ratings in MegaAcceptability v1.0 with two linear regressions---implemented in \texttt{scikit-learn} \citep{scikit-learn-2011}. We then compose these two regressions to map the normalized ratings in our extended MegaAcceptability dataset to those in MegaAcceptability v1.0. This gives us a combined dataset of acceptability judgments for sentences in three different tense/aspect combinations (\textit{past, present}, and \textit{past progressive}) and 50 different syntactic frames, which we use to construct our neg-raising experiment.

\section{Validation Experiments}
\label{sec:validation}

We conduct experiments aimed at validating our method for measuring neg-raising. In both experiments, we test the same set of 32 clause-embedding verbs, half of which we expect to show neg-raising behavior and the other half we do not (based on the literature discussed in \S\ref{sec:background}). For neg-raising verbs, we refer to the neg-raising predicates listed in \citealt{gajewski-2007} and \citealt{collins-postal-2018}; and for non-neg-raising verbs, we choose factive verbs and those that \citet{theiler-roelofsen-aloni-2017} claim are not neg-raising. The experiments differ with respect to whether we employ ``bleached'' items (as in the data collection described in the main body of the paper) or ``contentful'' items, which are constructed based on sentences drawn from English corpora. 

\paragraph{Materials}

We select neg-raising and non-neg-raising verbs such that half of each type takes infinitival subordinate clauses and half takes finite subordinate clauses. Table \ref{tab:pilot-verbs} shows the 32 verbs we choose for the pilot. Some verbs listed as taking one kind of subordinate clause can also take the other. In these cases, we only test that verb in the subordinate clause listed in Table \ref{tab:pilot-verbs}.
\begin{table}[t!]
\small
\begin{center}
\begin{tabular}{ p{1.4cm} p{2.3cm} p{2.2cm} }
\toprule
\textbf{Subordinate clause}   & \textbf{Neg-raising} & \textbf{Non-neg-raising} \\
\midrule
\textit{Finite} & think, believe, \par feel, reckon, \par figure, guess, \par suppose, imagine & announce, claim, \par assert, report, \par know, realize, \par notice, find out \\
\midrule
\textit{Infinitival} & want, wish, \par happen, seem, \par plan, intend, \par mean, turn out & love, hate, \par need, continue, \par try, like, \par desire, decide \\
\bottomrule
\end{tabular}
\caption{Verbs used in validation experiments}
\label{tab:pilot-verbs}
\vspace{-8mm}
\end{center}
\end{table}

The matrix subject (first v. third person) and matrix tense (present v. past) are manipulated for each predicate: \ref{ex:pilot-context2-bleached} schematizes four items from our bleached experiment and \ref{ex:pilot-context2-contentful} schematizes four items from our contentful experiment.

\ex.\label{ex:pilot-context2-bleached}
\{I, A particular person\} \{don't/doesn't, didn't\} want to do a particular thing.

\ex.\label{ex:pilot-context2-contentful}
\{I, Stephen\} \{don't/doesn't, didn't\} want to introduce new rules.

Items for the bleached experiment are constructed automatically using the templates, which select \textit{to have a particular thing} for \textit{turn out} and \textit{seem} as their subordinate clause, \textit{to do a particular thing} for other verbs taking infinitival subordinate clauses, and \textit{that something happened} for the verbs taking finite subordiante clauses. Items for the contentful experiment are constructed by replacing all bleached words (\textit{a particular person, a particular thing, do, have}, and \textit{happen}) from the bleached experiment items by contentful lexical words. 

The high content sentences are constructed based on sentences sampled from the Corpus of Contemporary American English \citep{coca-2017} and the Oxford English Corpus \citep{sketch-2014}. The contentful items are modified so that third person subject is a proper name and sentences do not include any pauses or conjunctions. To allow possible item variability, we create five contentful items per each bleached item.

For the bleached experiment, four lists of 32 items each are constructed by partitioning the resulting 128 items under the constraints that (i) every list contains every verb with exactly one subject (\textit{first}, \textit{third}) and tense (\textit{past}, \textit{present}) and (ii) every subject-tense pair is seen an equal number of times across verbs. We ensure that the same level of a particular factor is never assigned to the same verb more than once in any list and that the items in a list are randomly shuffled. To construct items, we manipulate neg-raising, embedded complement, matrix subject, matrix tense. Neg-raising and embedded complements are pre-determined for each verb, while matrix subject and matrix tense are randomly selected for a verb in each task. The same constraints apply for the contentful experiment except that the test items were partitioned into 20 lists of 32 instead of four lists because the total number of sentences for the contentful experiment is five times bigger than the bleached experiment. 

\paragraph{Participants}
For the bleached experiment, 100 participants were recruited such that each of the four lists was rated by 25 unique participants. For the contentful experiment, 100 participants were recruited as well, to have each of the 20 lists of 32 rated by five unique participants. No participant was allowed to rate more than one list. In each experiment, one participant out of 100 reported not speaking American English natively and this participant's responses were filtered prior to analysis.

\paragraph{Analysis}

We test whether our task correctly captures canonical (non-)neg-raising verbs using linear mixed effects models. For both validation experiments, we start with a model containing fixed effects for \textsc{negraising} (\textit{true}, \textit{false}; as in Table \ref{tab:pilot-verbs}), random intercepts for \textsc{participant}, \textsc{verb}, and (in the contentful validation) \textsc{item}. Nested under both verb and participant, we also included random intercepts for \textsc{matrix subject} (\textit{1st}, \textit{3rd}) and \textsc{matrix tense} (\textit{past}, \textit{present}) and their interaction. We compare this against a model with the same random effects structure but no effect of \textsc{negraising}. We find a reliably positive effect of \textsc{negraising} for both the bleached experiment ($\chi^2(1)=34.5$, $p<10^{-3}$) and the contentful experiment ($\chi^2(1)=19.8$, $p<10^{-3}$). This suggests that participants' responses are consistent with neg-raising inferences being more likely with verbs that have previously been claimed to give rise to such inferences.

\section{Normalization}
\label{sec:normalization}

For the purposes of visualization in \S\ref{sec:data}, we present normalized neg-raising scores. These scores are derived using a mixed effects robust regression with loss the same loss \ref{ex:finalloss} as for the model described in Section \ref{sec:model}, except that, unlike for the model, where $\nu_{vfjk}$ is defined in terms of the model, for the purposes of normalization, both $\nu_{vfjk}$ in \ref{ex:nrev} and $\alpha_{vfjk}$ in \ref{ex:accev} are directly optimized. Figure \ref{fig:neg-raising-result} plots $\mathrm{logit}^{-1}(\exp(\sigma_0)\nu_{vfjk}) + \beta_0$.

\section{Model Derivation}
\label{sec:modelappendix}

{\small
\begin{align*}
\mathbb{P}(d_{vf}) &= \mathbb{P}\left(\bigvee_t \lambda_{vt} \land \pi_{tf}\right)\\
&= \mathbb{P}\left(\lnot\lnot\bigvee_t \lambda_{vt} \land \pi_{tf}\right)\\
&= \mathbb{P}\left(\lnot\bigwedge_t \lnot(\lambda_{vt} \land \pi_{tf})\right)\\
&= \mathbb{P}\left(\lnot\bigwedge_t \lnot(\lambda_{vt} \land \pi_{tf})\right)\\
&= 1-\mathbb{P}\left(\bigwedge_t \lnot(\lambda_{vt} \land \pi_{tf})\right)\\
&= 1-\prod_t\mathbb{P}\left(\lnot(\lambda_{vt} \land \pi_{tf})\right)\\
&= 1-\prod_t 1-\mathbb{P}\left(\lambda_{vt} \land \pi_{tf}\right)\\
&= 1-\prod_t 1-\mathbb{P}(\lambda_{vt})\mathbb{P}(\pi_{tf})\\
\end{align*}
}

\end{document}